\documentclass[conference]{IEEEtran}
\usepackage{cite}
\usepackage{amsmath,amssymb,amsfonts}
\usepackage{algorithmic}
\usepackage{graphicx}
\usepackage{textcomp}
\usepackage{xcolor}
\usepackage{tabularx}
\usepackage{subcaption}
\usepackage{enumitem}

\usepackage{colortbl}
\usepackage{url}
\usepackage{hyperref}
\def\BibTeX{{\rm B\kern-.05em{\sc i\kern-.025em b}\kern-.08em
    T\kern-.1667em\lower.7ex\hbox{E}\kern-.125emX}}
\begin{document}

\title{Patient-Centric Knowledge Graphs: A Survey of Current Methods, Challenges, and Applications
}

\author{\IEEEauthorblockN{Hassan S. Al Khatib, Subash Neupane, Harish Kumar Manchukonda, Noorbakhsh Amiri Golilarz, \\Sudip Mittal, Amin Amirlatifi, Shahram Rahimi} Dept. of Computer Science \&  Engineering, Mississippi State University, Starkville, MS\\Email: \{hsa78, sn922\}@msstate.edu, \{amin\}@che.msstate.edu,\{amiri,mittal, rahimi\}@cse.msstate.edu\\
}

\maketitle

\begin{abstract}

Patient-Centric Knowledge Graphs (PCKGs) represent an important shift in healthcare that focuses on individualized patient care by mapping the patient's health information in a holistic and multi-dimensional way. PCKGs integrate various types of health data to provide healthcare professionals with a comprehensive understanding of a patient's health, enabling more personalized and effective care. This literature review explores the methodologies, challenges, and opportunities associated with PCKGs, focusing on their role in integrating disparate healthcare data and enhancing patient care through a unified health perspective. In addition, this review also discusses the complexities of PCKG development, including ontology design, data integration techniques, knowledge extraction, and structured representation of knowledge. It highlights advanced techniques such as reasoning, semantic search, and inference mechanisms essential in constructing and evaluating PCKGs for actionable healthcare insights. We further explore the practical applications of PCKGs in personalized medicine, emphasizing their significance in improving disease prediction and formulating effective treatment plans. Overall, this review provides a foundational perspective on the current state-of-the-art and best practices of PCKGs, guiding future research and applications in this dynamic field.

%\tiny
% \keyFont{ \section{Keywords:} Knowledge Graph, Patient-Centric, Personalized Healthcare, Natural Language Processing, Generative AI}
\end{abstract}


\section{Introduction} 
\label{introduction}
The healthcare industry has experienced a significant shift, transitioning from traditional, provider-centric models towards patient-centered care. This shift highlights the critical role of engaging patients as active participants in their healthcare journeys. At the center of this transformation are PCKGs, which significantly advance personalized, data-driven care. PCKGs facilitate the integration of diverse data types, including medical history, genetics, lifestyle choices, and real-time data from health technology devices, fostering a comprehensive view of patient health essential for customizing treatments to individual needs\cite{Mesko2022}\cite{Blobel2007}.

The evolution of Knowledge Graphs (KG) in healthcare into sophisticated networks reflects an increasing acknowledgment of the complexity of human health and the insufficiency of siloed data systems in addressing multifaceted health issues. This evolution is propelled by the necessity for a holistic understanding of patient health, enabling personalized care and applying advanced data analytics to enhance healthcare outcomes\cite{Blobel2011}\cite{shirai2021applying}. PCKGs stand at the forefront of healthcare innovation, signifying a crucial step towards integrated, patient-focused knowledge networks. This shift from fragmented data systems to cohesive KGs enables healthcare providers to employ machine learning and analytical technologies to help improve precision medicine, diagnostic accuracy, and treatment efficacy. Such a transition represents a technological leap that aligns with the broader objectives of healthcare reform aimed at delivering more personalized, efficient, and patient-centered care\cite{Albannai2019A}\cite{Almunawar2012}. However, incorporating KGs into healthcare presents technical, methodological, and ethical challenges, including data interoperability, privacy concerns, and the complexities of modeling diverse health outcomes. These hurdles pose significant barriers to the widespread adoption of PCKGs, yet the potential of these systems to revolutionize healthcare by offering a nuanced and comprehensive understanding of patient health is unquestionable\cite{MacLean2021Knowledge}\cite{Alagar2017}. 

PCKGs' primary aim is to enhance the quality of patient care, improve treatment outcomes, and increase the efficiency of healthcare delivery. By integrating disparate data sources and utilizing advanced analytical models, PCKGs promise to deliver personalized, efficient, and effective healthcare services tailored to each patient's unique needs. This goal emphasizes the shift towards a healthcare system that values and prioritizes individual patient experiences and needs, marking the beginning of a new era of patient-centric, data-driven care\cite{Harper2015}\cite{Goniewicz2021Integrated}. Despite the inherent challenges in integrating KGs into healthcare, the critical need for the advanced application of PCKGs to achieve personalized care and enhance healthcare delivery systems is undeniable. As the healthcare landscape continues to evolve, PCKGs exemplify the industry's commitment to leveraging technology to meet patients' complex and varied needs, thus marking a significant milestone in the journey toward more personalized and effective healthcare solutions.

The motivation of this paper stems from the growing need to consolidate disparate healthcare data into a unified, holistic view for improved patient care. The key contributions of this survey paper are:

\begin{itemize}
   \item A foundational explanation of knowledge graphs, serving as the theoretical basis for the remainder of the paper.
    \item Presentation of survey findings and introduction of a taxonomy developed for the field of PCKGs.
    \item An in-depth review of methodologies specifically designed for PCKGs, shedding light on the most effective techniques currently employed.
    \item Exploration of real-world applications and use-cases that have successfully implemented PCKG methodologies, providing evidence of their utility.
\end{itemize}

The rest of the paper is organized as follows: Section \ref{background} explains the principles of knowledge graphs, which provide a foundation for the discussions that follow. Section \ref{survey_approach} presents the findings from our survey and introduces the taxonomy we have developed. Moving on to Section \ref{current_methods}, we review methodologies explicitly designed for PCKGs. Section \ref{application} explores real-world applications and examples that benefit from these methodologies. Section \ref{current_challenges} highlights research challenges and provides targeted recommendations for future scholars in this field. Finally, in Section \ref{conclusion}, we summarize the key-findings of this paper and outline directions for work in PCKGs.

\section{Background}
\label{background}
The development of knowledge representation has a rich history in the realms of logic and AI. The notion of graphical knowledge representation can be traced back to 1956 when Richens \cite{richens1956preprogramming} introduced the concept of semantic nets. Similarly, symbolic logic knowledge finds its roots in the General Problem Solver \cite{newell1959report} of 1959. Initially, knowledge bases were employed in knowledge-based systems for reasoning and problem-solving. Notably, MYCIN \cite{shortliffe2012computer}, an expert system renowned for medical diagnosis, utilized a knowledge base containing approximately 600 rules. However, it was in 2012 that the concept of Knowledge Graph (KG) gained immense popularity, thanks to Google's search engine and its introduction of the Knowledge Vault framework \cite{dong2014knowledge}. This framework aimed to construct large-scale KGs through knowledge fusion. There is currently no consensus on the definition of the term, with several authors proposing different definitions. Table \ref{table:definition_kg} illustrates some of these definitions currently available in the literature.
\begin{table*}[ht]
{\renewcommand{\arraystretch}{1.30}%

\caption{Various definition of KGs in available literature.}

%\begin{tabularx}{\textwidth}{X|X|X|X|X|X|X}
%\begin{tabularx}{\textwidth}{X|X|X|X|X|X}
\begin{tabular}{p{2.4cm}|p{14.6cm}}

%\begin{tabular}{p{3.4cm}|l|l}
\hline
\rowcolor{cyan!20!}
Source         & Definition \\                                                                                          \hline

Paulheim et al. \cite{paulheim2017knowledge} & A knowledge graph (i) mainly describes real world entities and their interrelations, organized
in a graph, (ii) defines possible classes and relations of entities in a schema, (iii) allows for potentially interrelating arbitrary entities with each other and (iv) covers various topical
domains.\\ \hline
                                  
Ehrlinger et al. \cite{ehrlinger2016towards} & A knowledge graph
acquires and integrates information into an ontology and applies
a reasoner to derive new knowledge. \\ \hline

Wang et al. \cite{wang2017knowledge} & A knowledge graph is a multi-
relational graph composed of entities and relations which are
regarded as nodes and different types of edges, respectively. \\ \hline

Smirnova et al. \cite{Smirnova2018} & A Knowledge Graph, also known as a Knowledge Base, is a directed graph formed by triples that connect nodes (subjects) to other nodes (objects) through properties (predicates), where these connections represent semantic relationships and the graph's structure illustrates the subjects, objects, and their interrelations. \\ \hline

Duan et al. \cite{Duan2017} & A knowledge graph organizes items, entities, and users as nodes interconnected by edges, providing rich semantics and comprehensive information in a structure akin to natural language. \\ \hline

\end{tabular}
\\
\\

\label{table:definition_kg}
\vspace{-5mm}}

\end{table*}

\begin{figure}[h]
  \centering
  \includegraphics[width=0.75\linewidth]{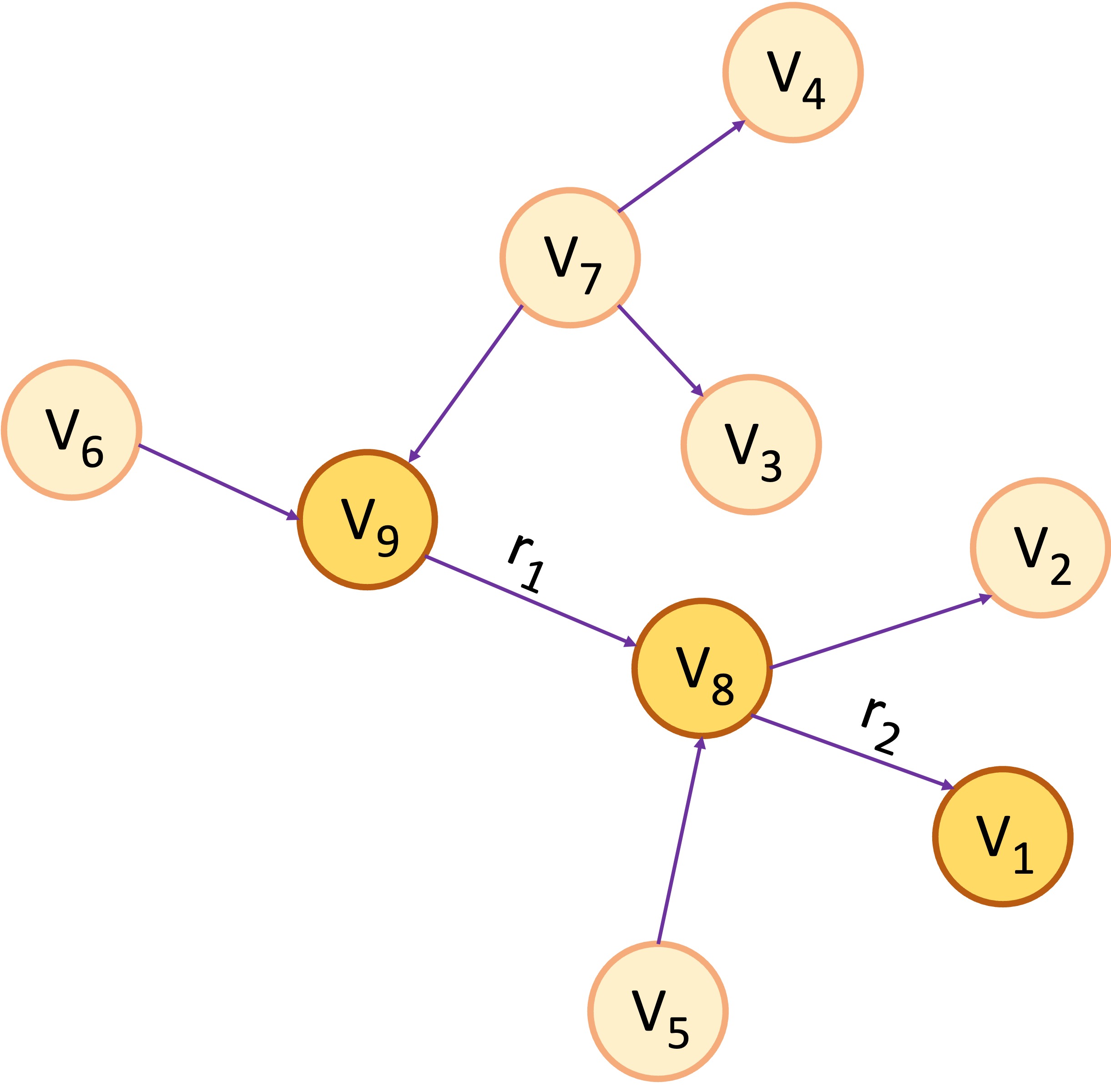}
  \caption{An example of knowledge graph, where the triplet ($v_9$, $r_1$, $v_8$) serves as an illustration of the link between entities $v_9$ and $v_8$ through the relation $r_1$ and ($v_8$, $r_2$, $v_1$) through $r_2$ for relation between $v_8$ and $v_1$.}
  \label{fig:rdf_kg}
\end{figure}

\begin{figure*}[htb]
  \centering
  \includegraphics[width=0.75\linewidth]{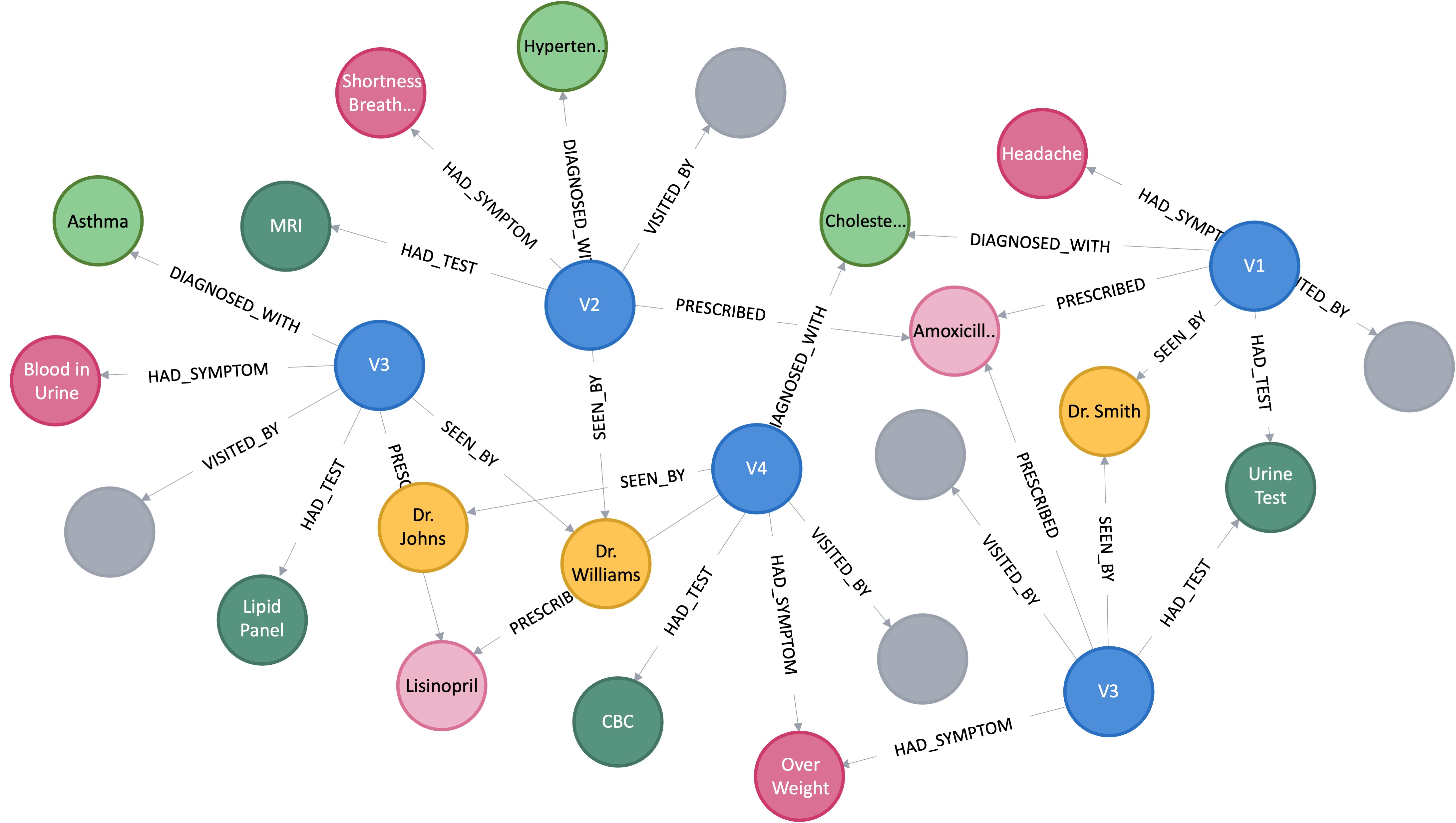}
  \caption{Illustration of Patient's Clincal Visits Knowledge Graph}
  \label{fig:visits}
\end{figure*}

A KG assumes the form of a directed graph $(G)$, characterized by vertices and edges, where, $G = (V,E)$. Vertices $(V)$ indicate an entity and Edge $(E)$ between the two vertices conveys the semantic relationship between two entities \cite{bordes2011learning}. Within the graph, vertices, also referred to as entities or nodes, are linked through relationships represented as edges and facts are depicted using RDF \cite{rdf} triples such as (subject, predicate, object) or (head, relation, tail), denoted by $<h,r,t>$ \cite{abu2021domain}.   Figure \ref{fig:rdf_kg} depicts a simple KG with vertex $v_9$ and $v_8$  linked by the relation $r_1$, which goes from $v_9$ to $v_8$, making a triplet ($v_9$, $r_1$, $v_8$). In this scenario, $v_9$ represents the head and $v_8$ represents the tail. In the context of Patient Centric Knowledge Graphs(PCKGs) nodes typically represent entities like patients, drugs, diseases, or genes, whereas edges denote relationships or associations between these entities (see Fig \ref{fig:visits}).

Over the past few years, especially within the domains of health and biomedicine, various forms of KGs have been proposed, often stemming from literature or Electronic Health Record (EHR), however they are typically not individual-centered. Recent work pertaining to disease-centric knowledge graphs are available in \cite{chandak2023building, lin2023multimodal}. These knowledge graphs facilitate clinicians' ability to study the relationships between diseases to answer practical clinical problems. Nevertheless, because relationships between disease concepts are frequently distributed across numerous datasets, the concept of integrating multimodal data sources is critical for developing comprehensive disease knowledge graphs \cite{chandak2023building, lin2023multimodal}. Their robustness of knowledge in terms of diseases and symptoms, on the other hand, must be evaluated in order to assess its accuracy and quality \cite{chen2019robustly}.

In hospitals, patient information collected during visits, as well as details of test results, clinical notes, symptoms, diagnoses, and drugs, is frequently maintained as an EHR. These records include both unstructured data in the form of free notes, such as progress notes, admission notes, discharge summaries, medical histories, procedures notes, etc., and structured data, such as medical codes, medications, administrative data, vital signs, and laboratory test report data \cite{sarwar2022secondary}. Numerous KGs have been introduced in the literature, utilizing EHRs as their foundation. As an illustration, Finlayson SG et al. \cite{finlayson2014building} constructed a graph including diseases, drugs, procedures and devices. To do so, they leveraged 1 million clinical concepts from 20 million clinical notes. Similarly, another study in \cite{rotmensch2017learning}, formulated a KG covering 156 diseases and 491 symptoms, based on the insights drawn from a dataset of 273,174 patient visits to the emergency department. Although some efforts have been made to incorporate patient-specific information into these graphs, they are not yet patient-centric.

PCKGs can be linked Personal Knowledge Graphs (PKG), an emerging concept in this field. Although the notion of PKGs is relatively new, there have been some notable contributions in this area. One of the pioneering papers that formally introduces the concept of PKG is authored by Balog et al.\cite{balog2019personal}. They define PKG  \textit{``to be a source of structured knowledge about entities and the relation between them, where the entities and the relations between them are of personal, rather than general, importance. The graph has a particular “spiderweb” layout, where every node in the graph is connected to one central node: the user''}. Within the scope of this definition, the PKG only incorporate knowledge pertinent to the specific user.

The emergence of KGs has brought about a substantial transformation in the healthcare domain, providing numerous advantages to this industry. Integrating disparate health data to provide a comprehensive and holistic view of patient health is a significant application of KGs. In the era of precision medicine, wherein therapeutic strategies are customized to suit individual patients' distinctive genetic composition and lifestyle variables, KGs assume a crucial function. Integrating various data types, such as genetic profiles, medical history, lifestyle behaviors, and information from health technology devices, is facilitated to enhance efficiency. Implementing this all-encompassing methodology enables more accurate diagnoses, treatment strategies, and potentially enhanced health results \cite{JohnKB-2020}.

Moreover, KGs are a formidable instrument for deciphering intricate biomedical data, resulting in amplified research findings \cite{BauerMA-2010}. They provide a systematic depiction of connections among different entities, such as genes, diseases, drugs, and pathways, which can be utilized to formulate novel hypotheses and uncover concealed patterns. For example, KGs  can elucidate the intricate relationship between a genetic mutation and its association with a specific disease or unveil the influence of environmental factors in the progression of said disease. As mentioned earlier, the discoveries possess the potential to significantly contribute to the development of groundbreaking therapeutic approaches and expand the horizons of biomedical investigation \cite{YuanY-2011}.

KGs  play a crucial role in developing effective decision support systems in the clinical domain. By effectively amalgamating and analyzing individualized patient data, KGs can assist medical practitioners in formulating accurate diagnostic determinations, devising optimal treatment strategies, and forecasting patient prognoses. Consequently, the implementation of KGs can effectively mitigate the occurrence of medical errors \cite{RaghW-2014}.

KGs have found significant applications across various sectors beyond healthcare. For example, in e-commerce, KGs are extensively used to enhance user experience through personalized recommendations, search result ranking, and semantic search. Prominent e-commerce platforms, such as Walmart, Amazon and eBay, have been utilizing KGs to link items to their respective attributes, thereby providing more relevant product suggestions to customers \cite{XuD-2020}.
Moreover, KGs are increasingly used in Natural Language Processing (NLP), where they can improve the performance of machine translation, question-answering, and text summarization systems \cite{Li2022ImprovingQA}. KGs also have substantial potential in cybersecurity, as they can aid in mapping relationships between cyber threats, vulnerabilities, and affected systems, facilitating better threat prediction and prevention \cite{Liu2022ARO}.
Social networking platforms like Meta and LinkedIn also leverage KGs to understand user relationships and interests, enhancing content recommendation and advertising targeting strategies \cite{HeQ-2020}. Hence, the applications of KGs are diverse and significantly impact various domains, reinforcing their relevance and potential.

\section{Developed Taxonomy of Patient-Centric Knowledge Graph}
\label{survey_approach}
Having introduced the concept of KG in the previous section, we now explore the survey's core. This section will examine the various approaches, categorizations, and details that influence PCKGs. 
PCKGs are characterized by a variety of methodologies and studies. In order to navigate this diverse landscape, we need a well-structured framework. In this context, a taxonomy is a guide that enables readers to comprehend and classify the complexities involved in PCKG construction, evaluation, processing, and applications.

\begin{figure*}[htb]
  \centering
  \includegraphics[width=1.0\linewidth]{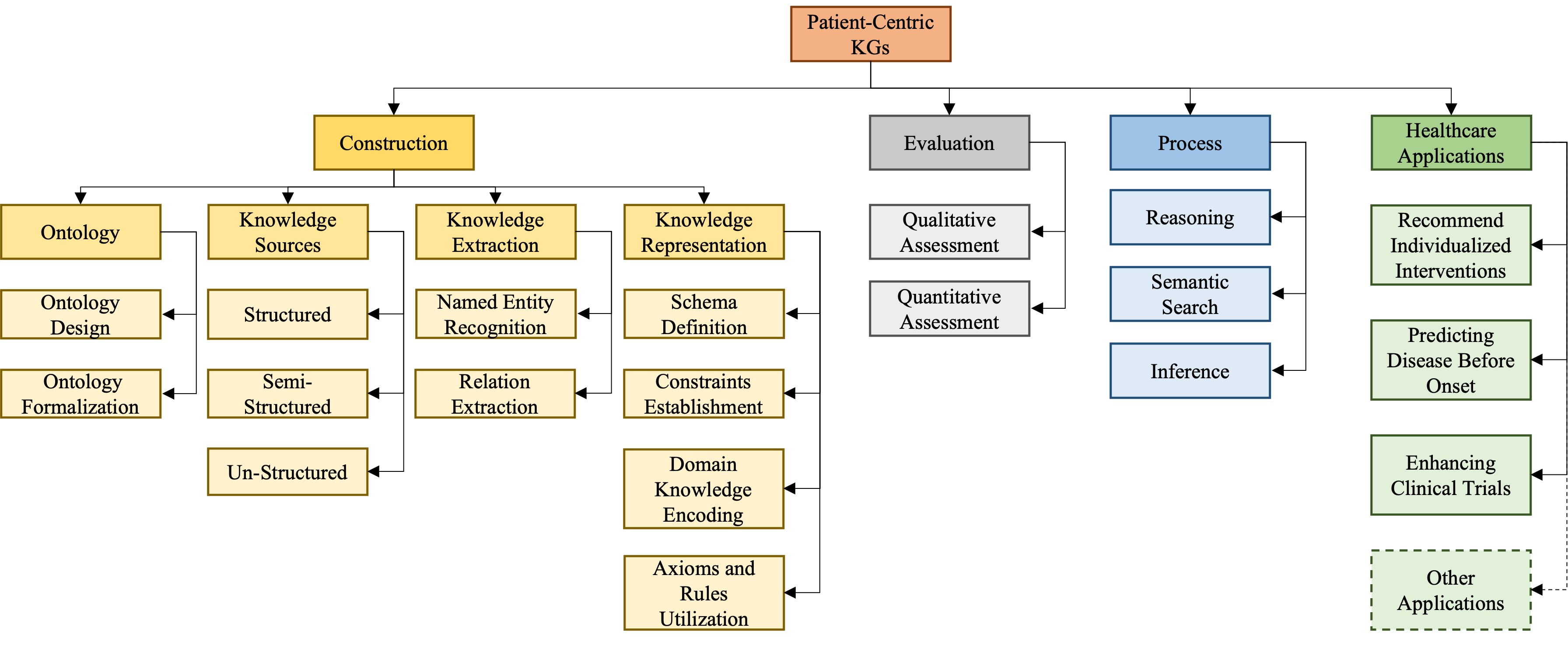}
  \caption{Proposed Taxonomy of Patient-Centric Knowledge Graph}
  \label{fig:taxonomy}
\end{figure*}

Our developed taxonomy as depicted in Fig \ref{fig:taxonomy}
is derived from an exhaustive survey and analysis of existing literature and practices.
Based on our findings, we classify PCKG into four major category including \textit{construction, evaluation, process and applications}. Given the heterogeneity of studies in the domain, it provides a structured representation and insight into the why and how of each categorization. The taxonomy presented in this paper outlines essential relationships and highlights the important aspects of each of these categories. Its uniqueness lies in its exhaustive nature, informed by the foundational literature and cutting-edge research.
In the following paragraphs, we delve into each of the categories and subcategories.

The first category pertains to \textit{construction} of a PCKG. It is further divided into four sub-categories including \textit{ontology, knowledge sources, knowledge extraction, and knowledge representation.}

 The initial step, \textit{Ontology}, revolves around designing structured frameworks that are custom-tailored to meet patient-centered needs, ensuring that the graph accurately captures the essence of patient data while also adhering to industry standards and best practices. It is important to articulate distinctions among closely related terms in this context: a Knowledge Graph (KG), which represents interconnected real-world facts; a Knowledge Base (KB), which stores this structured information; and an ontology, which provides a structured framework that illustrates and organizes concepts, ensuring consistent understanding and interpretation of the data within the KG and KB. Second, considering the diverse nature of healthcare data, our \textit{Knowledge Sources} segment highlights the differences and integration methods for three main sources of data: structured, semi-structured, and unstructured data. Third, the \textit{Knowledge Extraction} phase employs techniques like Named Entity Recognition (NER) to pinpoint essential entities such as drugs or symptoms. The process is complemented by Relationship Extraction (RE), which accurately maps the relationships among these entities, giving the graph its characteristic interconnected structure. The last sub-category in construction is \textit{Knowledge Representation}, which focuses on detailing the methodological approach of encapsulating both explicit and implicit knowledge in the PCKG. This involves defining the schema for entities and relationships, establishing constraints, and encoding domain knowledge through axioms and rules, thus constructing a robust framework that effectively translates raw data into a comprehensible and interconnected knowledge graph that accurately mirrors the complexity and depth of patient-centric data.

The second category involves the evaluation of the PCKG, emphasizing the need to assess the efficacy and accuracy of PCKGs post-construction. This assessment can be facilitated by two main methodologies such as \textit{qualitative assessment} and \textit{quantitative assessment} . The former involves a thorough examination of the graph to ensure that it adheres to clinical best practices and contains relevant and correct medical relationships. While the latter makes use of statistical and computational methodologies to evaluate the graph's performance metrics, including accuracy, recall, and precision, and offers a numerical assessment of its ability to accurately and effectively represent and infer knowledge. Once the PCKG has been constructed and evaluated, the next step is its \textit{utilization (processing PCKG)} to derive actionable insights. To attain this objective, various approaches are crucial, including the utilization of methods such as \textit{reasoning, semantic search, and inference}. \textit{Reasoning} enables the PCKG to logically extract new information and knowledge from the existing knowledge within the graph, providing a basis for making intelligent decisions and predictions about healthcare outcomes and strategies. While, \textit{Semantic Search} allows for structured searches to retrieve precise information from the graph. \textit{Inference}, on the other hand, uses the inherent structure of the graph to derive new insights and conclusions.

The final category of the our taxonomy relates to \textit{application} of PCKG in healthcare domain. We explore four primary healthcare use cases, encompassing \textit{recommending individualized interventions, clinical trials, predicting disease before onset, and others.} The first one personalizes patient care treatments based on insights extracted directly from the graph. While the second application aims to enhance the efficiency of \textit{Clinical Trials}, aiding in developing trial designs or optimizing patient selection. On the other hand, a particularly innovative application is \textit{Predicting Disease Before Onset}, where the complex patterns and relationships mapped in the graph can be harnessed to proactively detect potential health issues. Although these are highlighted applications, the potential of PCKGs extends even further, covering areas such as \textit{drug discovery, personalized medication regimens,} and \textit{preventive health strategies}, to name a few.

Throughout the next sections, we will examine each component of our taxonomy in more detail in order to gain a better understanding of its significance as well as provide an overview of the suggested implementation steps.

\vspace{10pt}

\section{Construction, Evaluation and Processing of PCKGs}
\label{current_methods}
Throughout this section, we explore the complex processes and methodologies involved in developing, assessing, and utilizing PCKG. This comprehensive exploration is structured into four critical categories as illustrated in Fig \ref{fig:taxonomy}.
The first category, \textit{construction} of PCKG, explains the complex process of building PCKG, highlighting the significance of ontology, various data sources, knowledge extraction techniques, and knowledge representation. Next, \textit{evaluation} covers both qualitative and quantitative methods for assessing the efficacy and accuracy of these graphs. Similarly, we discuss advanced techniques like \textit{reasoning, semantic search}, and \textit{inference}, which are essential to harnessing the full potential of PCKGs to derive actionable insights and inform healthcare decisions. 

\subsection{PCKG Construction}
The process of creating PCKGs involves four steps. To begin with, the \textit{ontology} phase involves developing structured frameworks based on patient data, which ensure accuracy and adherence to standards. \textit{Knowledge sources} is the second step, which addresses the collection and integration of diverse types of healthcare data, including structured, semi-structured, and unstructured data. During the third step, \textit{knowledge extraction}, key data entities and their interconnections are mapped using named entity recognition and relationship extraction techniques. Lastly, \textit{knowledge epresentation} involves defining a schema for entities and relationships, setting constraints, and encoding domain knowledge to transform raw data into a coherent, interconnected graph reflecting patient data complexity. In the following subsections, we will explore each of these steps in detail.

\subsubsection{Ontology}
Ontologies provide a structured framework for representing knowledge, enabling the integration of diverse data sources and facilitating sophisticated access to health information, which is crucial in patient-centered healthcare \cite{puustjarvi2010providing}. The design and formalization of ontologies involve creating a comprehensive \textit{schema} that integrates information from various \textit{structured} and \textit{unstructured} data sources \cite{iannacone2015developing}. Ontologies are different from knowledge bases and schemas. While an ontology is a formal representation of a set of concepts within a domain and the relationships between those concepts, a knowledge base is a collection of knowledge in a specific area, often built upon an ontology. A schema, on the other hand, is more about the organization of data as a blueprint of how the database is constructed, without delving into the domain-specific relationships and properties that an ontology provides. The application of ontologies in healthcare has been demonstrated in the body of literature. For instance, Sun et al. \cite{sun2014information} presented an ontology-enabled healthcare service provision model that facilitates joint referral decisions between patients and general practitioners, promoting patient-centric healthcare service provision. The following subsections will expand more on the design and formalization of ontologies.

\begin{figure}[htb]
  \centering
  \includegraphics[width=0.90\linewidth]{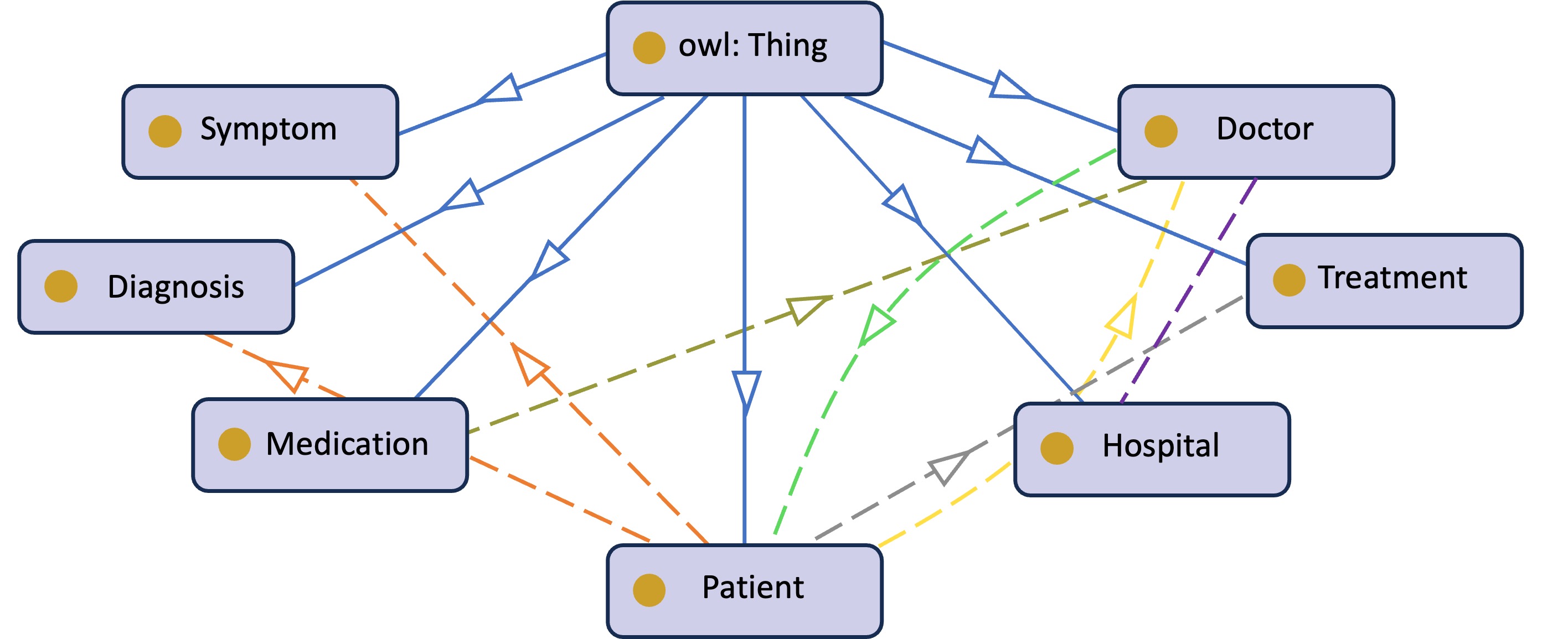}
  \caption{Illustration of a Basic Patient's Ontology}
  \label{fig:ontology}
\end{figure}

\paragraph{Design}
The growing digitization of healthcare data and the proliferation of medical information need effective and structured methods of representing and managing patient-centric knowledge. Ontologies as formal representations of knowledge, play a critical role in supporting interoperability, data integration, and decision support \cite{TayeM-2010},  in healthcare systems. In the context of PCKGs, ontology design is a fundamental process that involves defining and modeling the important entities, connections, and features related to patient health and treatment. Figure \ref{fig:ontology} 
provides an example of a patient's ontology. Authors in \cite{ontology} argue that the rationale behind ontology development is to aid in a shared understanding of information structure among people or software agents, facilitate the reuse of domain knowledge, make domain assumptions explicit, separate domain knowledge from operational knowledge, and analyze domain knowledge. However, there is not standard methodology \cite{ kapoor2010comparative} or no single correct way of formulating ontology \cite{ontology}. 

The initial stage in ontology design is to establish the scope and domain of the PCKG. The scope of the ontology specifies the breadth of medical knowledge included, whereas the domain defines the precise areas of concentration, such as general medical information, specific diseases, or healthcare procedures. Identifying the domain aids in the selection of acceptable current ontologies such as SNOMED-CT \cite{elsappagh2018snomed}, RxNorm \cite{hanna2013building}, and LOINC \cite{uchegbu_potential_2017} or healthcare standards that can be utilized to maintain interoperability and consistency. While current ontologies and standards may provide a solid foundation, some patient-centric concepts may not be fully covered. In such circumstances, additional classes and features must be added to cater to specific requirements. This requires collaboration between domain experts and ontology engineers to guarantee that the ontological representations are consistent with healthcare practices and guidelines.
    
The second step is to define the core entities. Entities such as \emph{Patient}, \emph{Disease}, \emph{Medication},\emph{Symptom}, \emph{Medical Procedure},  \emph{Healthcare Provider}, etc. are typically included. These entities serve as the building blocks for constructing the KG and are crucial for capturing patient-related information.
    
Next step is to extract relationship or specifying relationship between entities. This step is important for capturing the complex interactions within patient-centric healthcare. Relationships, such as \emph{Diagnosis, Treats, Has Symptom, and Undergoes,} enable meaningful associations and support inferencing capabilities. The developer of ontology during this stage should also carefully consider associated attributes and properties of the entities. For example, attributes of \emph{patient} could be \emph{age}, \emph{gender}, \emph{sex}, \emph{demography} etc.  These attributes can influence the diagnosis and treatment decisions \cite{bravo2019methodology}.

\paragraph{Formalization}
Formalizing the ontology focuses on structuring and representing patient data and medical knowledge in a way that is understandable and accessible to healthcare systems. Puustjarvi et al. \cite{puustjarvi2010providing} leveraged the Resource Description Framework (RDF) and Web Ontology Language (OWL), to facilitate sophisticated access to health information, a cornerstone for effective patient care. This innovative approach underscores the evolving landscape of healthcare informatics, where the structuring and accessibility of patient data are extremely important.Similarly, Jiang et al. \cite{jiang2003context} provided insight into the role of Formal Concept Analysis (FCA) in the development of ontologies within clinical domains. FCA emerges as a key tool, offering linguistic and context-based knowledge. This knowledge is indispensable for clinical experts, aiding them in comprehending and applying ontology effectively in their practice. Integrating FCA into ontology development signifies a deeper understanding of clinical data and its nuances, enhancing the overall utility of these ontologies in real-world medical settings.

Further broadening this scope, Djedidi and Aufaure \cite{djedidi2007medical} presented a methodology that underscores the integration of heterogeneous data sources with existing ontologies and standards. This approach is instrumental in constructing a medical domain ontology as a comprehensive, knowledge-centric decision support system. Such integration is crucial in ensuring the developed ontology is robust and aligns seamlessly with existing medical knowledge frameworks and data sources. Likewise, Mendes, Rodrigues, and Baeta \cite{mendes2013development} made significant contributions to the development of the Ontology for General Clinical Practice (OGCP). This ontology extends the Ontology for General Medical Science (OGMS) by incorporating the Clinical Patient Record (CPR) structure. This extension significantly enhances the representation and reasoning capabilities within clinical practice knowledge, especially in the context of natural language text. The OGCP stands as a testament to the evolving complexity and sophistication required in modern medical ontologies, catering to the nuanced needs of clinical practice.

The dynamic nature of medical knowledge and patient data necessitates methodologies for tracking and updating information. In this context, Ognyanov and Kiryakov \cite{ognyanov2002tracking} proposed a formal model for tracking changes in RDF(S) repositories. This model is vital for maintaining the accuracy and relevance of the KG over time, addressing the ever-changing landscape of medical data and knowledge. Addressing the user perspective, Clarkson et al. \cite{clarkson2018usercentric} introduced a user-centric methodology for the ontology population. This approach aligns user concepts with target ontologies, proving an efficient method for building and maintaining ontologies across various domains. The user-centric approach ensures that the developed ontologies are technically sound and resonate with the needs and understandings of the end-users, be they patients or healthcare professionals.

To bridge the gap between formal ontology and practical application, Ferilli \cite{ferilli2021integration} proposed an intermediate format that can be easily mapped onto formal ontology for complex reasoning and a graph database for efficient data handling. This innovation represents a significant stride in harmonizing the theoretical aspects of ontology with the practical demands of data management, ensuring that the developed ontologies are both conceptually sound and practically applicable.

Together, these diverse yet interconnected research efforts paint a comprehensive picture of ontology formalization's current state and future potential in patient-centric healthcare. They highlight a collaborative and multi-faceted approach to developing technically robust ontologies that are deeply integrated with the practical realities of healthcare delivery.

\subsubsection{Knowledge Sources}

\begin{figure}[htb]
  \centering
  \includegraphics[width=0.90\linewidth]{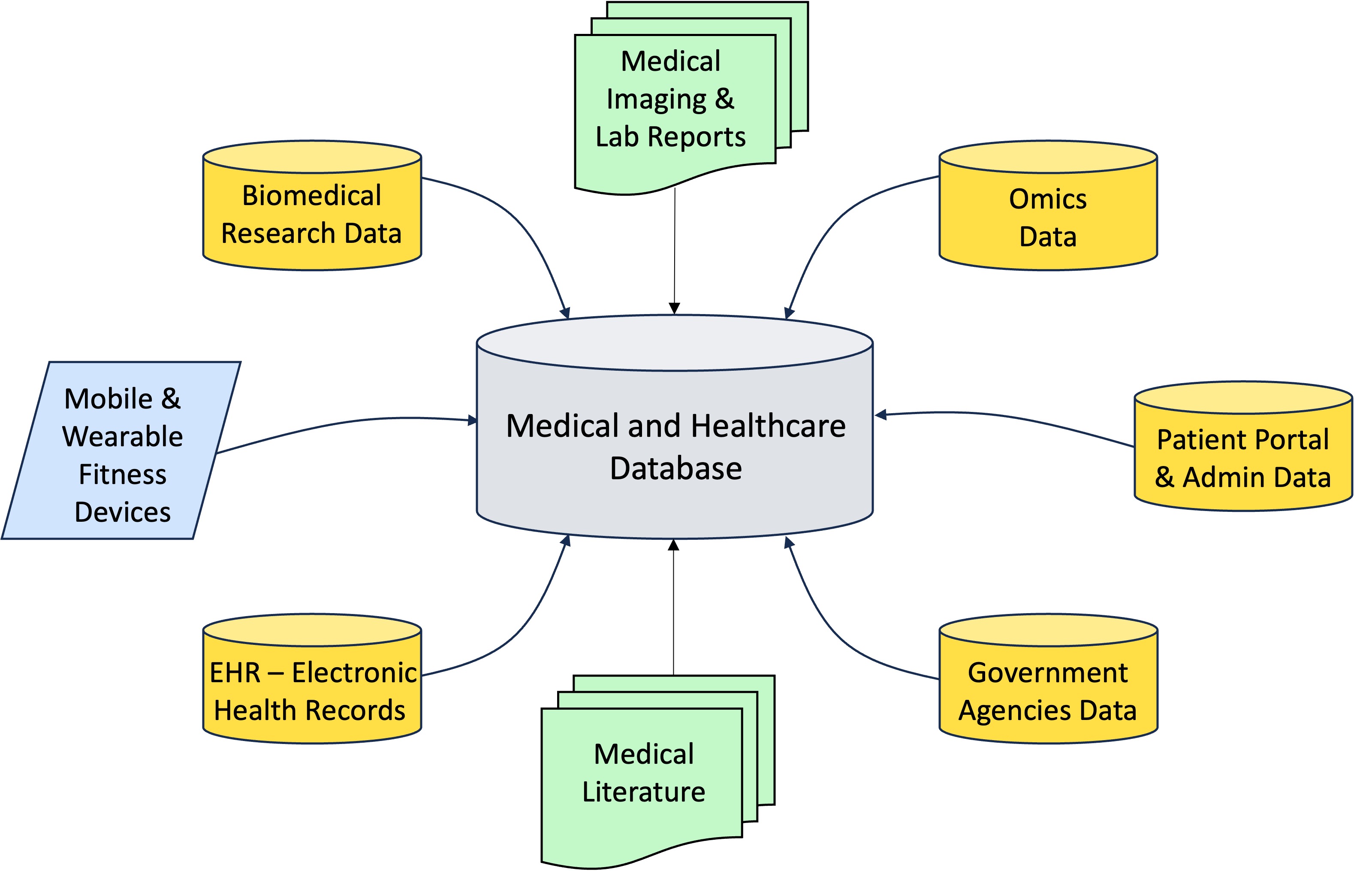}
 \caption{Diverse Sources of Medical and Healthcare Data}
  \label{fig:data}
\end{figure}

PCKGs are complex structures that integrate diverse data sources, as illustrated in Fig \ref{fig:data},
in order to offer a comprehensive perspective on a patient's medical past, present ailments, and prospective therapies. The extent of the depth of a PCKG is contingent upon the diversity and quality of the data sources that contribute to its composition. The sources can be classified into three main categories: structured, semi-structured, and unstructured data.

\paragraph{Structured Data}
Structured data is organized and easily searchable in relational databases. It follows a specific schema or model, making it straightforward to query and analyze. Structured healthcare data, integral to PCKGs, is exemplified by \textit{EHRs, lab results and dagnostics}, and \textit{genomic and molecular data}. \textit{EHRs}, as digital records of a patient's medical history, encompass diagnoses, medications, treatment plans, and vital statistics, providing a highly structured backbone for PCKGs. Transitioning to \textit{lab results and diagnostics}, this category includes structured, numerical, or categorical data from blood tests, imaging studies (like X-rays, MRIs), and other diagnostic tests, facilitating their integration into PCKGs. Lastly, \textit{genomic and molecular data}, derived from genetic tests, offer structured insights into a patient’s disease predisposition, proving essential for the advancement of personalized medicine. Collectively, these structured data forms are key in healthcare analytics and personalized treatment planning.

\paragraph{Semi-Structured Data}
Semi-structured data does not reside in a relational database but does have some organizational properties that make it easier to analyze. It often requires some level of transformation to be fully utilized. Semi-Structured healthcare data encompasses\textit{clinical notes} and \textit{patient-generated data}, each possessing unique characteristics that bridge the gap between structured and unstructured data. \textit{Clinical notes and narratives}, while fundamentally unstructured in text, often adhere to templates or contain tags, rendering them semi-structured and providing contextual depth absent in purely structured datasets. Similarly, \textit{patient-generated data}, sourced from wearables like fitness trackers and mobile health apps, presents structured data points (e.g., heart rate, steps) but lacks a consistent schema overall, contributing to the semi-structured nature of the data set. This combination of structured elements within an otherwise unstructured framework is valuable for comprehensive healthcare data analysis.

\paragraph{Unstructured Data}
Unstructured data is information that doesn't have a pre-defined data model, making it challenging to analyze using conventional methods. Unstructured healthcare data, characterized by its non-standardized format, includes \textit{patient surveys} and \textit{interviews} as well as \textit{medical literature and research papers}, each offering unique challenges and insights for PCKGs. \textit{Patient surveys and interviews}, typically in text or audio formats, yield qualitative insights into patients' conditions but are unstructured, necessitating advanced analytics for effective integration into PCKGs. Similarly, \textit{medical literature and research papers} are a rich source of valuable insights but remain unstructured; thus, they require NLP techniques to distill and extract relevant information for PCKG incorporation. This unstructured data, with its inherent complexity, plays a crucial role in enhancing the depth and breadth of healthcare analytics and patient care understanding.

By integrating these diverse data sources, PCKGs can offer a multi-dimensional view of a patient's health, thereby enabling more personalized and effective healthcare interventions.

\subsubsection{Knowledge Extraction}

In the process of constructing PCKGs, knowledge extraction serves a pivotal role. This stage leverages techniques such as Named Entity Recognition (NER) and Relation Extraction (RE) to distill valuable insights from unstructured data. While NER is focused on identifying and categorizing key entities present in the data, RE takes a step further to identify the relationships that exist between these entities. Through the harmonious interplay of NER and RE, a structured and rich KG can be constructed, paving the way for more personalized and efficient healthcare solutions.

\paragraph{Named Entity Recognition (NER)}
NER involves the identification and classification of entities such as diseases, symptoms, and medications from unstructured text. This process is fundamental in transforming raw data into structured, actionable information. Recent advancements in NER methodologies have significantly enhanced the development of KGs, particularly in the medical domain. For instance, Li et al. \cite{li2019attention} introduced the BiLSTM-Att-CRF model, which integrates an attention mechanism and part-of-speech features to improve clinical NER in Chinese electronic medical records. Building on this, Keretna, Lim, and Creighton \cite{keretna2015enhancement} proposed a graph-based technique that enhances medical NER performance by up to 26\% in unstructured medical text, showcasing the versatility of NER applications.

Furthering these developments, Wang and He \cite{wang2021bi} developed a bi-directional joint embedding model that combines encyclopedic knowledge with original text, thereby showing improved results in Chinese medical NER. Complementing this approach, Wang, Xia, Zhou, Ruan, Gao, and He \cite{wang2018incorporating} demonstrated that incorporating dictionaries into deep neural networks effectively addresses the challenge of rare and unseen entities in clinical NER, thus broadening the scope of NER applications. In a similar vein, Guan and Tezuka \cite{guan2022medical} highlighted the importance of entity linking and intent recognition in medical question-answering systems, which significantly improves the efficiency and accuracy of medical KG searches. In a similar context, Zhou, Cai, Zhang, Guo, and Yuan \cite{zhou2021mtaal} developed a multi-task adversarial active learning model that enhances medical NER and normalization by considering task-specific features, further illustrating the dynamic evolution of NER methodologies.

These diverse methodologies reflect the evolving landscape of NER in the development of PCKGs. The integration of advanced machine learning techniques, attention mechanisms, and domain-specific knowledge has led to more accurate and efficient entity recognition, which is crucial for the effective utilization of KGs in healthcare.

\paragraph{Relation Extraction (RE)}
Similar to NER, RE is a fundamental process in transforming unstructured data into a structured form, enabling more effective data utilization in healthcare and medical research. Recent advancements in RE methodologies, particularly in the context of PCKGs, have been significant. Ruan et al. \cite{ruan2020relation} demonstrated that a multi-view graph learning method could notably enhance the precision, recall, and F1-score in relation extraction from Chinese clinical records, outperforming state-of-the-art methods. Similarly, Li et al. \cite{li2019survey} highlighted the evolution of relation extraction algorithms, emphasizing the role of deep learning, reinforcement learning, active learning, and transfer learning in this domain.

In the realm of therapy-disease KGs, Wang et al. \cite{wang2021construction} constructed a Therapy-Disease KG (TDKG) using entity relationship extraction, achieving an 88.98\% accuracy in extracting valid relationships from treatment-disease literature. This underscores the potential of RE in enhancing the accuracy and comprehensiveness of medical KGs. Furthermore, the integration of KGs and lexical features in biomedical relation extraction, as shown by Zhao et al. \cite{zhao2020knowledge}, indicates significant improvements in semantic understanding. This approach exemplifies the innovative strategies adopted in recent research to enhance the effectiveness of RE in medical contexts. The development of Conditional
Random Field (CRF)-powered classification models with deep learning for clinical relation extraction, as demonstrated by Lv et al. \cite{lv2016clinical}, further illustrates the ongoing advancements in this field. These models have shown effectiveness in extracting clinical relation information from medical texts, thereby improving the construction of medical KGs.

The field of RE in PCKGs is rapidly evolving, with methodologies leveraging multi-view graph learning, deep learning, and knowledge-enhanced approaches. These advancements are crucial in extracting meaningful and accurate relationships from complex medical texts, thereby enriching the KGs that are essential for advancing patient-centered care and medical research.

\subsubsection{Knowledge Representation}
Knowledge Representation (KR) serves as a methodological backbone, enabling the systematic transformation of both explicit and implicit knowledge into a structured and interpretable format. This transformation is crucial for machines to effectively process, analyze, and infer from the data, thereby bridging the gap between raw data and actionable insights. 

The process of KR encompasses four key components such as schema definition, constraint establishment, domain knowledge encoding and \textit{axioms and rules utilization}  as explained in Section \ref{survey_approach}. Firstly, \textit{schema definition} involves outlining the structure of entities and their relationships within the KG. This step is fundamental in organizing data in a meaningful and accessible manner. Secondly, \textit{constraint establishment} is essential for maintaining data integrity and consistency, ensuring that the relationships and entities adhere to predefined rules and norms. Thirdly, incorporating \textit{domain knowledge} is instrumental in embedding domain-specific intelligence into the system. This allows for a nuanced understanding of the specific area of interest. Fourthly, the integration of \textit{axioms and rules} facilitates advanced reasoning capabilities, enabling the system to infer new knowledge and insights based on the existing data.

These components collectively form a robust framework that simplifies complex data and enhances the system's capability to mirror the intricacies inherent in patient-centric data. In the following subsections, we will delve deeper into each of these components, exploring their roles and significance in the broader context of KR.

\paragraph{Schema Definition}

The concept of \textit{schema definition} in PCKGs revolves around structuring and organizing patient-related data in a way that enhances understanding and interaction with this information. Schemas in KGs serve as blueprints for organizing data, enabling more effective data integration, querying, and analysis. This is particularly crucial in healthcare, where patient data is complex and multifaceted. It aims to create a unified and comprehensive view of patient information, facilitating better healthcare outcomes through informed decision-making. Ghosh and Gilboa \cite{ghosh2014memory} emphasize the role of memory schemas in coordinating knowledge structures, highlighting the importance of understanding cognitive processes in schema development. This perspective is crucial in PCKGs, where patient data must be organized in a way that aligns with healthcare professionals' cognitive schemas.

Building on the concept of visual schemas by Kranjec et al.\cite{kranjec2013schemas}, which enhance comprehension and aid in the diagnostic process, the work of Ohira et al. \cite{ohira2011introducing} takes a step further by introducing the dynamic nature of graph-based data models. This flexibility is key in the realm of PCKGs, where patient data is not static but continuously evolving. The ability to dynamically add and remove relationships in these KGs ensures that they remain up-to-date and reflective of the latest patient information, a critical aspect in the fast-paced environment of healthcare. Transitioning from the dynamic structuring of patient data, the focus shifts to the cognitive aspects of PCKGs as discussed by Gilboa and Marlatte \cite{gilboa2017neurobiology}. Their research underscores the significance of aligning PCKGs with the mental models of healthcare professionals. This alignment is not just about structuring data efficiently; it's about ensuring that the KG resonates with the users' cognitive processes, thereby enhancing memory recall and decision-making capabilities.

Further advancing this discussion, Ji et al. \cite{ji2018survey} delve into the intricacies of schema induction in KGs. Their exploration into the challenges and developments in this field is particularly relevant for PCKGs. As these KGs become more complex and integral to healthcare, understanding and overcoming the challenges in schema induction is paramount for ensuring that PCKGs are robust, comprehensive, and seamlessly integrated into the healthcare workflow.

\paragraph{Constraints Establishment}
The establishment of constraints in PCKGs focuses on enhancing the accuracy, privacy, and efficiency of medical data management and analysis. Various methodologies and strategies adopted in the creation and application of constraints in PCKGs, as evidenced by recent academic research. For example, Aguilar-Escobar et al. \cite{aguilar2016applying} demonstrate the application of the Theory of Constraints (TOC) in improving the logistics of medical records in hospitals, highlighting the potential of TOC in enhancing service quality and reducing costs in healthcare settings. This approach underscores the importance of efficient data management in PCKGs.

Building on this notion of efficiency, the focus shifts to the critical aspect of privacy and security in healthcare, where Mathew and Obradovic \cite{mathew2011constraint} illustrate how constraint graphs can secure privacy in medical transactions, preventing unauthorized attribute-based transformations in clinical data. This is crucial in PCKGs, where patient data sensitivity is paramount. The evolution of PCKGs further extends into the domain of dynamic data handling and adaptability. Matsuto and Shiina \cite{matsuto1990laboratory} propose a model for medical knowledge representation based on constraints, which dynamically generates laboratory schedules using constraint propagation. This approach emphasizes the adaptability of PCKGs in accommodating patient-specific information.

The adaptability of PCKGs is further enhanced by advancements in KG reasoning. Wu et al. \cite{wu2017knowledge} discuss a constraint-based embedding model for KG reasoning, focusing on semantic-type constraints in constructing corrupted triplets. This methodology significantly improves the reasoning accuracy of KGs, a key aspect in developing effective PCKGs. Alongside reasoning accuracy, the representation of knowledge and uncertainties also plays an important role. Pugh and Gillan \cite{pugh2020propositional} introduce Propositional Constraint Graphs (PCG) as a tool for representing knowledge and uncertainties in various tasks, including healthcare. This approach aids in the clear visualization and management of complex information in PCKGs.

The application of these concepts is not limited to static scenarios but also extends to dynamic and evolving medical challenges. Tu \cite{tu2021constraint} proposes a constraint propagation approach for identifying biological pathways in COVID-19 KGs. This method, utilizing semantically extracted information, is indicative of the potential for PCKGs in pandemic response and other rapidly evolving medical scenarios. Finally, the integration of context-aware constraints brings a new dimension to the integrity and validation of knowledge in PCKGs. Wilcke et al. \cite{wilcke2020bottom} discuss the use of context-aware constraints to improve knowledge integrity in heterogeneous KGs. This balance of complexity and utility is crucial for knowledge validation tasks in PCKGs \cite{wilcke2020bottom}, highlighting the continuous evolution and application of constraints in the development of robust and efficient PCKGs.

The establishment of constraints in PCKGs is a multifaceted process, involving the integration of various methodologies to enhance data accuracy, privacy, and efficiency. The strategies adopted in recent research works reflect a growing emphasis on adaptability, security, and precision in managing patient-centric data in healthcare systems.

\paragraph{Domain Knowledge Encoding}
Domain Knowledge encoding focuses on the integration and representation of complex medical knowledge in a structured and interconnected format. This encoding process is crucial for developing PCKGs, which are instrumental in enhancing patient care through personalized and informed decision-making. The methodologies for creating domain knowledge encoding in PCKGs are diverse and innovative. Wang et al. \cite{wang2016diagnosis} proposed a framework utilizing medical chart and note data, employing a bag-of-words encoding method and a model that considers both global information and local correlations between diseases. This approach underscores the importance of capturing the nuances of medical data for effective KG construction.

Further expanding on these methodologies, Ji et al. \cite{ji2020survey} highlighted the significance of representation space, scoring function, encoding models, and auxiliary information in KG creation. This comprehensive approach indicates the multi-faceted nature of KG development, where various components contribute to the robustness and accuracy of the resulting graph. Similarly, Yuan et al. \cite{yuan2019constructing} presented an approach for biomedical domain KG construction that includes entity recognition, unsupervised entity and relation embedding, latent relation generation, clustering, relation refinement, and relation assignment. This method demonstrates the complexity involved in accurately representing medical knowledge in graph form. Rotmensch et al. \cite{rotmensch2017learning} on the other hand, utilized rudimentary concept extraction and three probabilistic models to construct high-quality health KGs, with the Noisy OR model yielding the best results. This study exemplifies the use of probabilistic models in enhancing the quality of KGs. Furthermore, Yu et al. \cite{yu2020relationship} proposed a relationship extraction method for domain KG construction, obtaining upper and lower relationships from structured data, semi-structured data, and unstructured text. This method highlights the importance of extracting relationships from diverse data sources to enrich the KG.

These methodologies contribute to the construction of comprehensive and accurate KGs. In addition, they also pave the way for innovative applications in patient-centric healthcare.

\paragraph{Axioms and Rules Utilization}
PCKGs utilize axioms and rules to enhance the representation and analysis of patient data, leading to more informed healthcare decisions. Axioms in PCKGs are fundamental principles or statements accepted as true without proof, used to define relationships and properties within the graph. Rules, on the other hand, are logical statements that infer new knowledge from existing data within the graph. This combination of axioms and rules plays an important role in structuring and interpreting complex medical data, enabling more personalized and effective patient care.

The methodologies for creating axioms and rules in PCKGs vary across different research works. Chu et al. \cite{chu2021knowledge} demonstrated the adaptation of multi-center clinical datasets by incorporating an external KG, which enhanced patient features and improved predictions for acute kidney injury in heart failure patients. This approach signifies the importance of integrating diverse data sources and knowledge bases into PCKGs. In another study, Tomczak and Gonczarek \cite{tomczak2012decision} developed a Graph-based Rules Inducer for extracting decision rules from data streams in diabetes treatment. This method supported medical interviews by tracking hidden context changes and avoiding overfitting, highlighting the importance of dynamic rule adaptation in response to evolving data. Shang et al. \cite{shang2021ehr} utilized an EHR-oriented KG system to effectively harness non-used information buried in EHRs, thereby improving healthcare quality and providing interpretable recommendations for specialist physicians.

The use of graph-based association rules for mining medical databases, as demonstrated by Alzoubi \cite{alzoubi2013mining}, revealed the potential of discovering hidden medical knowledge, which could contribute significantly to understanding complex medical conditions like preterm birth. Similarly, RuleHub, as introduced by Ahmadi et al. \cite{ahmadi2020rulehub}, is an extensible corpus of rules for public KGs, enabling users to archive and retrieve rules from popular KGs. This tool improves data understanding and reduces redundant work, illustrating the importance of rule sharing and standardization in the field. Wright et al. \cite{wright2011method} developed a set of rules for inferring patient problems from clinical and billing data, which performed better than using problem list or billing data alone. This method improved clinical decision support and care improvement, showcasing the practical application of rule-based systems in clinical settings.

Based on our review, the utilization of axioms and rules in PCKGs is a dynamic and evolving field, with methodologies ranging from the integration of diverse data sources to the development of specialized, disease-centric graphs. These approaches enhance the understanding of complex medical data and significantly contribute to the advancement of personalized patient care.

\vspace{10pt}

\subsection{PCKG Evaluation}

Evaluation of PCKGs is critical for ensuring their efficacy and accuracy in representing and inferring medical knowledge. These evaluations are typically conducted using two main methodologies: \textit{qualitative} and \textit{quantitative} assessments. Qualitative assessment involves a detailed examination of the KG to ensure it aligns with clinical best practices and accurately reflects medical relationships. This method may usability studies, content analysis, and expert panel reviews to gauge the graph's relevance and correctness\cite{rotmensch2017learning} \cite{Brundage2015}. On the other hand, quantitative assessment employs statistical and computational techniques to measure performance metrics such as accuracy, recall, and precision, providing a numerical evaluation of the graph's effectiveness\cite{rotmensch2017learning}.

\subsubsection{Quantitative Assessment} Quantitative methodologies for PCKG evaluation focus on numerical metrics to assess the graph's performance. These methods include \textit{completeness, consistency, accuracy, and embedding techniques}.

Completeness in a KG refers to the extent to which all necessary information is represented. In healthcare, this means ensuring that a PCKG includes comprehensive data on diseases, symptoms, treatments, and patient histories. A systematic literature review by Issa et al. \cite{issa2021knowledge} emphasizes the importance of assessing the completeness of KGs, identifying various methodologies and metrics used for this purpose. On the other hand, consistency involves checking for logical coherence within the graph, particularly in terms of medical terminologies and relationships. This ensures that the KG does not contain contradictory information, which is crucial for clinical decision-making. The work of Sharma and Bhatt \cite{sharma2022privacy} on privacy-preserving KGs in healthcare highlights the importance of maintaining consistency in data representation.

Other methods like accuracy assessment and embedding techniques are integral to the development of reliable PCKGs. The accuracy of a KG, as highlighted by Li et al. \cite{li2020realworld}, is crucial for patient safety and effective treatment planning, ensuring that the information aligns with authoritative medical databases and literature. Complementing this, embedding techniques like node2vec or GraphSAGE, as demonstrated by Talukder et al. \cite{talukder2022diseasomics}, play a vital role in transforming complex graph data into a more accessible format, thereby enhancing the graph's ability to accurately capture and represent semantic relationships in healthcare applications.

\subsubsection{Qualitative Assessment} Qualitative assessment methodologies are instrumental in evaluating the effectiveness of PCKGs in meeting the needs of their intended users, including clinicians, researchers, and patients.

In the qualitative assessment of PCKGs, usability studies and feedback loops play an important role in the KG evaluation. Usability studies, as highlighted by Brundage et al. \cite{Brundage2015}, are crucial for evaluating the ease of use and accessibility of PCKGs, particularly in the context of interpreting patient-reported outcomes (PROs) in graphic format. These studies typically employ a blend of qualitative and quantitative methods, including surveys, interviews, and user testing, to gather comprehensive feedback on the user experience. Complementing this, feedback loops, as discussed by Rospocher et al. \cite{rospocher2016building}, enable users to provide direct input on their experiences with the KG. This feedback is instrumental for the continuous refinement and adaptation of the KG, ensuring it aligns closely with user needs and preferences.

Furthermore, the quality of PCKGs is also gauged through comparison with established databases and expert validation. Benchmarking PCKGs against renowned medical databases like PubMed and ClinicalTrials.gov is essential for assessing their content coverage and accuracy \cite{huang2017constructing}\cite{huang2017constructingdisease}. This comparison helps in identifying any gaps or discrepancies in the PCKG. Additionally, expert validation, underscored by the work of Rotmensch et al. \cite{rotmensch2017learning}, involves domain experts who ensure the medical relevance and accuracy of the PCKG's content. Their insights into the clinical applicability of the information within the KG are vital for maintaining its reliability and trustworthiness.

The evaluation of PCKGs through qualitative and quantitative assessments is a multifaceted process that involves both the review of medical content and the analysis of performance metrics. Medical professionals and patients have benefited from progress in this field by ensuring that PCKGs serve as reliable and effective tools.

\vspace{10pt}
\subsection{PCKG Processing}
By processing PCKGs, healthcare providers can extract actionable insights that inform clinical decision. The utilization of PCKGs involve different methods including reasoning, semantic search, and inference.

\subsubsection{Reasoning}
KG Reasoning in healthcare extends beyond mere data aggregation, playing an important role in deriving new insights and facilitating informed decision-making. This reasoning process involves structuring information, extracting features and relations, and performing logical deductions to uncover new knowledge from existing data within the graph \cite{Rajabi2022}. For instance, in the context of mental healthcare, KGs have been utilized for emotion recognition from facial expressions and heart rate, demonstrating the potential of knowledge reasoning in predicting emotional states \cite{Yu2020}. Moreover, KGs in healthcare support clinical decision-making and enhance hospital efficiency by integrating heterogeneous medical knowledge and services, thereby enabling cognitive computing and semantic reasoning \cite{Ding2021}.

\subsubsection{Semantic Search}
It enables the retrieval of specific information by structuring searches to navigate the complex relationships within the graph. Various strategies have been adopted to enhance the precision and efficiency of information retrieval in PCKGs. For instance, the use of Semantic Web and Knowledge Management approaches has been pivotal in implementing patient-centric strategies with well-defined semantics\cite{lytras2009semantic}. Graph-based methods that incorporate semantic-rich knowledge bases and lazy learning algorithms have shown promise in linking multimodal clinical data for improved diagnosis performance \cite{wang2018automatic}. Moreover, the retrieval of similar clinical cases has been refined by mapping text to Unified Medical Language System (UMLS) concepts and representing patient records as semantic graphs, demonstrating superiority over traditional models\cite{plaza2010retrieval}.
\subsubsection{Inference}
Inference uses the graph’s inherent structure to deduce new insights, facilitating the discovery of patterns and trends that may not be immediately apparent\cite{wang2018automatic}\cite{lytras2009semantic}. These methods underscore the advances in accuracy and efficiency in the processing and utilization of PCKGs, leading to enhanced healthcare outcomes.

After examining the methodologies for building, evaluating, and processing PCKGs, we move on to practical applications and use cases that will demonstrate how PCKGs can be applied in real-world scenarios, demonstrating their impact and significance in healthcare.

\section{Applications \& Use cases}
\label{application}
As the healthcare industry increasingly embraces data-driven decision-making, PCKGs have emerged as a powerful tool for personalized medicine. These KGs, which place the patient at the center of a complex network of interconnected health data, offer a holistic view of a patient's health journey. They encompass various data, from medical history and genetic information to lifestyle factors and environmental influences. The applications of these KGs are broad and innovative, offering the potential for improved disease prediction, enhanced treatment planning, and more effective preventive strategies. In the following subsections, we will explore these applications in detail, shedding light on how PCKGs revolutionize healthcare and pave the way for a more personalized, predictive, and proactive approach to patient care. 

\subsection{Predicting Disease Before Onset} 

The healthcare sector has seen significant advancements in patient-centric solutions by integrating KGs. Using a KG to proactively forecast the likelihood of a disease developing in an individual before any symptoms appear involves synthesizing various data points related to patient history, genetic information, lifestyle factors, and broader medical knowledge to identify patterns and risk factors associated with diseases. This predictive model allows for early intervention strategies, significantly altering the disease trajectory and improving the patients' long-term health outcomes.

The methodologies for creating PCKGs for disease prediction are diverse and innovative. Chen et al. \cite{chen2019robustly} discuss the robust extraction of medical knowledge from EHRs to build graphs that evaluate accuracy across different diseases and patient demographics using non-linear functions for causal relationships. Similarly, Talukder et al. \cite{talukder2022diseasomics} describe an AI-based approach integrating disease-related knowledge bodies with Node2VEC for link prediction in disease-symptom networks. Furthermore, Liang et al. \cite{liang2022deep} highlight the use of multi-hop reasoning over KGs, which provides interpretability and is superior to single-hop methods. In similar vein, the refinement of medical KGs using latent representations aids in prediction and maintains the explainability of diagnoses \cite{heilig2022refining}. Expanding on these advancements, Li et al. \cite{li2020graph} introduce a Graph Neural Network-Based Diagnosis Prediction (GNDP) model that uses spatial-temporal graph convolutional networks for diagnosis predictions. In a complementary manner, Zheng et al. \cite{zheng2021multi} proposed a Multi-modal Graph Learning framework (MMGL) that exploits multi-modality information for disease prediction.

Different strategies across research works include the use of random walk along KG \cite{sun2020interpretable}, Graph Neural Networks (GNN) for embedding medical concepts \cite{sun2020disease}, and hybrid systems combining KGs with clinical experience for pediatric disease prediction \cite{liu2016hkdp}. Zhang et. al \cite{zhang2021knowledge} built an automatic question-answering system based on medical KGs, while Saha et al. \cite{saha2020predicting} predicted missing and noisy links in clinical KGs using neighborhood-based embeddings. Key findings from our literature review indicate significant advances in disease prediction accuracy, efficiency, and outcomes. For instance, the integration of GNNs with EMR data has led to highly representative node embeddings that improve prediction accuracy \cite{sun2020disease} . The use of tensor factorization on biological KGs for predicting co-morbid disease pairs \cite{biswas2019relation} and the construction of KGs based on evidence-based medicine for diabetes complications \cite{wang2020construction} are notable advances.

PCKGs represent a significant step forward in the field of disease prediction. They offer the potential to transform patient outcomes by enabling early detection and personalized treatment plans. Future implications include the continued refinement of these models to enhance their predictive power and the integration of even more diverse data sources to capture the full spectrum of patient health and disease progression. In Table \ref{table:app_predict} we summarize a list of selected disease prediction applications in terms of their impact and limitations in the field of PCKGs.

\begin{table*}
{\renewcommand{\arraystretch}{1.30}

\caption{A summary of selected literature on using PCKGs for \textit{``Predicting Disease Before Onset''}}
\label{table:app_predict}
\centering

%{\renewcommand{\arraystretch}{1.30}%
 
    \begin{tabularx}{\textwidth}{|X|X|X|X|}
    \hline
    \rowcolor{lightgray!20!}
    \textbf{Paper Title} & 
    \textbf{Focus/Objective} & 
    \textbf{Contribution(s)}&
    \textbf{Limitation(s)}\\
    \hline
    %==========================================%
    \rowcolor{cyan!20!}
    \multicolumn{4}{c}{Predicting Disease Before Onset} \tabularnewline
    \hline
   %==========================================%
    A novel link prediction approach on clinical knowledge graphs utilizing graph structures \cite{dorpinghaus2022novel} &
    
    The goal is to create a proof of concept showcasing the efficacy of graph structures in AI methodologies. &
    
    \begin{minipage}[t]{\linewidth}
    \begin{itemize}[leftmargin=*]
         \item Development of a graph-based method merging Conditional Random Fields (CRFs) and graph embedding for knowledge discovery.
        \item This method successfully predicts labels for graph nodes with high precision and recall.
    \end{itemize} 
    \vspace{1mm}
    \end{minipage} &
    \begin{minipage}[t]{\linewidth}
    \begin{itemize}[leftmargin=*]
        \item The method is time-intensive when querying features from the graph, particularly with large datasets.
        \item Increased runtime and memory demands for multi-node paths in the graph.
    \end{itemize} 
    \vspace{1mm}
    \end{minipage} \\
  
    \hline 

    %==========================================%
Deep knowledge reasoning guided disease prediction \cite{liang2022deep} &
    
    The paper aims to enhance model interpretability by integrating knowledge entities with single-hop and multi-hop relationships &
    
    \begin{minipage}[t]{\linewidth}
    \begin{itemize}[leftmargin=*]
        \item Introduction of HitaNet, a hierarchical time-aware attention network.
        \item HitaNet uses a self-attention based transformer model for enhanced disease prediction.
        \item The paper presents a unique token wrapping method to merge knowledge graph insights with EHR data.
    \end{itemize} 
    \vspace{1mm}
    \end{minipage} &
    \begin{minipage}[t]{\linewidth}
    \begin{itemize}[leftmargin=*]
        \item Lower time efficiency compared to baseline methods as a limitation.
        \item The proposed method's effectiveness diminishes with the availability of sufficient data.
    \end{itemize} 
    \vspace{1mm}
    \end{minipage} \\
  
    \hline 

    %==========================================%
Predicting missing and noisy links via neighbourhood preserving graph embeddings in a clinical knowlegebase \cite{saha2020predicting} &
    
    The paper proposes a model that combines support vector classification (SVC) and neural network-based probabilistic embedding (NPE) to predict the links between clinical entities in the KG &
    
    \begin{minipage}[t]{\linewidth}
    \begin{itemize}[leftmargin=*]
        \item The development of a novel approach that integrates SVC and NPE for link prediction in KGs.
        \item The model achieved promising results in terms of predicting missing associations.
    \end{itemize} 
    \vspace{1mm}
    \end{minipage} &
    \begin{minipage}[t]{\linewidth}
    \begin{itemize}[leftmargin=*]
        \item The reliance on existing clinical databases, which may contain noisy or incomplete data.
        \item The potential mismatch between model predictions and clinical recommendations.
    \end{itemize} 
    \vspace{1mm}
    \end{minipage} \\
  
    \hline 
    
    %==========================================%
Relation prediction of co-morbid diseases using knowledge graph completion \cite{biswas2019relation} &
    
    The objective is to predict relationships between co-morbid diseases using knowledge graph completion techniques &
    
    \begin{minipage}[t]{\linewidth}
    \begin{itemize}[leftmargin=*]
        \item Proposing a method that combines distributed representation learning and graph embedding to predict disease-disease relationships
        \item Development of a novel approach that leverages the structure and semantics of a KG to predict disease relationships
    \end{itemize} 
    \vspace{1mm}
    \end{minipage} &
    \begin{minipage}[t]{\linewidth}
    \begin{itemize}[leftmargin=*]
        \item The proposed method relies on the availability of a comprehensive and accurate knowledge graph.
        \item The evaluation of the method is limited to a specific dataset.
    \end{itemize} 
    \vspace{1mm}
    \end{minipage} \\
  
    \hline 

    %==========================================%
Disease prediction via graph neural networks \cite{sun2020disease} &
    
    The paper addresses the challenges in predicting both common and rare diseases by integrating expert knowledge with machine learning techniques. &
    
    \begin{minipage}[t]{\linewidth}
    \begin{itemize}[leftmargin=*]
        \item Proposing a systematic solution that combines expert knowledge with machine learning.
        \item Introducing a novel graph embedding-based model for disease prediction, which learns embeddings from medical concept graphs and patient record graphs.
    \end{itemize} 
    \vspace{1mm}
    \end{minipage} &
    \begin{minipage}[t]{\linewidth}
    \begin{itemize}[leftmargin=*]
        \item Struggling to adapt to rare diseases due to data scarcity and complex symptom-diagnosis relationships.
        \item Reliance on historical patient records for model training limits the ability to serve new patients.

    \end{itemize} 
    \vspace{1mm}
    \end{minipage} \\
  
    \hline

    %==========================================%
    \end{tabularx}

}
\end{table*}

\subsection{Recommending Individualized Interventions}

Intervention recommendation aims to use KGs to suggest tailored medications and treatments. By mapping patients' unique health profiles, these graphs enable physicians to pinpoint optimal therapies that enhance the precision of medical decisions, leading to improved patient outcomes. The application of KGs in personalized treatments is predicated on the integration of diverse data sources, including EHRs, clinical notes, and patient-generated data, to construct a comprehensive, interconnected data structure that reflects individual patient profiles. This patient-centric approach is crucial as it allows for treatments to be tailored based on the unique medical history, genetic information, lifestyle, and preferences of each patient, which is a departure from the one-size-fits-all healthcare model.

The approaches to developing and implementing PCKGs differ among research studies, but they all aim to achieve robustness, accuracy, and personalization. For instance, Gyrard et al. \cite{gyrard2018personalized} aggregate knowledge from IoT devices, clinical notes, and EMRs to manage chronic diseases, showcasing the integration of AI and machine learning in constructing Personalized Healthcare Knowledge Graphs (PHKGs). Transitioning from a general approach to a more specialized application, Individualized Knowledge Graphs (IKGs) in cardiovascular medicine are one strategy for developing these KGs, which combine biological knowledge with medical histories and health outcomes to create personalized treatment strategies \cite{ping2017individualized}. Further refining the methodological framework, Rotmensch et al. \cite{rotmensch2017learning} utilized concept extraction and probabilistic models, finding the Noisy OR model particularly effective for constructing high-quality health KGs. Building on these methodologies, Shirai, et al. \cite{shirai2021applying} applied Personal Knowledge Graphs (PKGs) to integrate patient-specific information into decision-making tools for personalized healthcare. In a similar vein of enhancing disease treatment strategies through personalized data, Zhu et al. \cite{Zhu2022RD} introduced a KG that enhances rare disease (RD) treatment recommendations by systematically compiling and semantically annotating RD-related scientific articles, aggregating essential research findings and therapeutic insights with a sophisticated data model.

Literature has demonstrated significant advances in treatment accuracy, efficiency, and outcomes. For instance, the use of Four-Tuple Path Matrix in Traditional Chinese Medicine has been proposed to create personalized KGs, enhancing diagnostic modalities \cite{xie2018personalized}. Li et al. \cite{li2014personal} demonstrated how personal KGs could be automatically constructed from user utterances in conversational dialogs, indicating the potential for real-time, dynamic treatment adjustments. PCKGs provide multiple advantages. By analyzing patient profiles using KGs, treatments are more accurate and efficient than traditional methods. For example, Zhang et al. \cite{zhang2022developing} developed an intuitive graph representation of knowledge for nonpharmacological treatment of psychotic symptoms in dementia, which could potentially transform care strategies for such complex conditions.

There is a wide range of potential implications for future personalized treatment and patient outcomes. A more informed and dynamic approach to treatment has the potential to enhance medical precision, improve patient engagement, and optimize health outcomes by integrating PCKGs. It is expected that these methods will become more sophisticated as the field evolves, allowing for even greater personalization and efficacy in treatment. In Table \ref{table:app_treat} we summarize a list of selected treatment decision applications in terms of their impact and limitations in the field of PCKGs.

\begin{table*}
{\renewcommand{\arraystretch}{1.30}

\caption{A summary of selected literature on using PCKGs for \textit{``Recommending Individualized Intervention''}}
\label{table:app_treat}
\centering

%{\renewcommand{\arraystretch}{1.30}%
 
    \begin{tabularx}{\textwidth}{|X|X|X|X|}
    \hline
    \rowcolor{lightgray!20!}
    \textbf{Paper Title} & 
    \textbf{Focus/Objective} & 
    \textbf{Contribution(s)}&
    \textbf{Limitation(s)}\\
    \hline
    %==========================================%
    \rowcolor{cyan!20!}
    \multicolumn{4}{c}{Recommending Individualized Interventions} \tabularnewline
    \hline
   %==========================================%
    Individualized Knowledge Graph \cite{ping2017individualized} &
    \begin{minipage}[t]{\linewidth}
    \begin{itemize}[leftmargin=*]
        \item Envisioning individualized Knowledge Graphs (iKGs) in cardiovascular medicine.
        \item Proposing a modern informatics platform for transforming clinical and scientific discovery. 
    \end{itemize} 
    \vspace{1mm}
    \end{minipage}&
    
    \begin{minipage}[t]{\linewidth}
    \begin{itemize}[leftmargin=*]
         \item Introducing the concept of iKGs for aggregating and presenting individualized cardiovascular health data.
         \item Highlighting the role of iKGs in linking biological and clinical knowledge of individual patients.
    \end{itemize} 
    \vspace{1mm}
    \end{minipage} &
    \begin{minipage}[t]{\linewidth}
    \begin{itemize}[leftmargin=*]
        \item Acknowledging challenges in data fragmentation, noncommensurability, and semantic inference within cardiovascular data.
    \end{itemize} 
    \vspace{1mm}
    \end{minipage} \\
  
    \hline 

    %==========================================%
Personalized Health Knowledge Graph \cite{gyrard2018personalized} &
    
    Aims to manage chronic diseases more effectively using IoT data analytics and explicit knowledge.&
    
    \begin{minipage}[t]{\linewidth}
    \begin{itemize}[leftmargin=*]
        \item Proposes a methodology to build PHKG, integrating heterogeneous data sources.
        \item Offers a solution for contextualizing and personalizing healthcare information.
    \end{itemize} 
    \vspace{1mm}
    \end{minipage} &
    \begin{minipage}[t]{\linewidth}
    \begin{itemize}[leftmargin=*]
        \item The paper acknowledges the complexity in semantic integration of diverse data.
        \item It highlights the challenges in tailoring generic knowledge to individual patients.
    \end{itemize} 
    \vspace{1mm}
    \end{minipage} \\
  
    \hline 

    %==========================================%
Developing an intuitive graph representation of knowledge for nonpharmacological treatment of psychotic symptoms in dementia \cite{zhang2022developing} &
    
    \begin{minipage}[t]{\linewidth}
    \begin{itemize}[leftmargin=*]
        \item Develop a knowledge graph for nonpharmacological treatment of psychotic symptoms in dementia.
        \item Enhance understanding and management of dementia-related psychotic symptoms through nonpharmacological methods.
    \end{itemize} 
    \vspace{1mm}
    \end{minipage} &
    
    \begin{minipage}[t]{\linewidth}
    \begin{itemize}[leftmargin=*]
        \item Creation of the Dementia-Related Psychotic Symptom Nonpharmacological Treatment Ontology (DRPSNPTO).
        \item Improvement in visualization and computerization of gerontological knowledge.
    \end{itemize} 
    \vspace{1mm}
    \end{minipage} &
    \begin{minipage}[t]{\linewidth}
    N/A
    \vspace{1mm}
    \end{minipage} \\ 
  
    \hline

%==========================================%
Learning a health knowledge graph from electronic medical records \cite{rotmensch2017learning} &
    
Automatically learn a health KG from EMRs to link diseases and symptoms and improve clinical decision-support systems. &
    
    \begin{minipage}[t]{\linewidth}
    \begin{itemize}[leftmargin=*]
        \item A methodology for deriving health KG from EMR using probabilistic models.
        \item Demonstration that the noisy OR model significantly outperforms other tested models.
    \end{itemize} 
    \vspace{1mm}
    \end{minipage} &
    \begin{minipage}[t]{\linewidth}
    \begin{itemize}[leftmargin=*]
        \item Inherent difficulties in interpreting EMR data, especially the presence of complex patient conditions.
        \item The reliance on rudimentary concept extraction pipelines.
        \item Limitations related to the automatic inference of causal relationships from observational data.
    \end{itemize} 
    
    \vspace{1mm}
    \end{minipage} \\
  
    \hline 
    
%==========================================%
Applying personal knowledge graphs to health \cite{shirai2021applying} &
    
The paper focuses on leveraging PHKGs to enhance healthcare decision-making by integrating personal health information with broader knowledge graphs. &
    
    \begin{minipage}[t]{\linewidth}
    \begin{itemize}[leftmargin=*]
        \item Proposing a conceptual framework for PHKGs, highlighting how they can support personalized, knowledge-driven healthcare applications by leveraging data from EHRs, IoT devices, and other health-related data sources.
    \end{itemize} 
    \vspace{1mm}
    \end{minipage} &
    \begin{minipage}[t]{\linewidth}
    \begin{itemize}[leftmargin=*]
        \item Collecting and storing personal health knowledge from heterogeneous sources.
        \item Linking personal health knowledge to external KGs enhances the PHKG with broader contextual information.
        \item Maintaining the PHKG to ensure it remains up-to-date and accurate.

    \end{itemize} 
    
    \vspace{1mm}
    \end{minipage} \\
  
    \hline 
    %==========================================%
    \end{tabularx}

}
\end{table*}

\subsection{Enhancing Clinical Trials}

KGs represent a paradigm shift in clinical trail patient selection by offering a structured, interconnected data framework that can encapsulate complex patient information, medical histories, and potential trial criteria. In clinical trials, patient-centric approaches are crucial because personalized medicine tailors treatments to each patient's characteristics, necessitating a comprehensive understanding of patient data.

PCKGs are created and applied using diverse methodologies. Gortzis and Nikiforidis \cite{gortzis2008tracing} described an N-tier system that combines KGs with human collaboration and scalable knowledge engineering tactics. In order to select patients effectively, expert input must be combined with scalable data structures. Xiang et al. \cite{xiang2019knowledge} highlighted the standardization and structural integration provided by KGs, which are essential for auxiliary diagnosis systems in clinical trials. Strategies for developing KGs for patient selection in clinical trials include the linkage of multimodal data types for automatic diagnosis \cite{wang2018automatic}. Nicholson and Greene \cite{nicholson2020constructing} discussed machine learning methods for constructing low-dimensional representations of KGs, which support applications in genomic, pharmaceutical, and clinical domains.

Various studies have indicated significant improvements in the accuracy, efficiency, and outcomes of clinical trial patient selection. For instance, the Safe Medicine Recommendation (SMR) framework by Wang et al. \cite{wang2017safe} bridges electronic medical records with medical KGs to learn patient-disease-drug embeddings, enhancing the precision of clinical trial patient selection. Compared with traditional ways of selecting patients for clinical trails, PCKGs offer a more thorough and subtle approach. Several studies have demonstrated the effectiveness of KGs in improving clinical trial design and outcomes by providing a more holistic view of patient data, facilitating personalized trial matching, and providing a more holistic view of patient data \cite{sharma2015patient} \cite{weng2017framework} \cite{huang2017constructing}. Utlimately, the integration of patient-centric KGs in the selection of clinical trial participants has the potential to transform the field by improving the precision and personalization of patient care. As a result of a better understanding of patient data, future trials will likely be more adaptive, patients will be more engaged, and outcomes may be improved. In Table \ref{table:app_trial} we summarize a list of selected treatment decision applications in terms of their impact and limitations in the field of PCKGs.

\begin{table*}
{\renewcommand{\arraystretch}{1.30}

\caption{A summary of selected literature on using PCKGs for \textit{``Enhancing Clinical Trials''}}
\label{table:app_trial}
\centering

%{\renewcommand{\arraystretch}{1.30}%
 
    \begin{tabularx}{\textwidth}{|X|X|X|X|}
    \hline
    \rowcolor{lightgray!20!}
    \textbf{Paper Title} & 
    \textbf{Focus/Objective} & 
    \textbf{Contribution(s)}&
    \textbf{Limitation(s)}\\
    \hline
    %==========================================%
    \rowcolor{cyan!20!}
    \multicolumn{4}{c}{Recommending Individualized Interventions} \tabularnewline
    \hline
   %==========================================%
Knowledge graph-based clinical decision support system reasoning \cite{xiang2019knowledge} &
    
    The focus of the paper is to highlight the benefits of using knowledge graphs over traditional hand-crafted rule databases in CDSSs &
    
    \begin{minipage}[t]{\linewidth}
    The introduction of the Path-Ranking Algorithm (PRA) as a method for automatically discovering symptoms without human intervention.
    \vspace{1mm}
    \end{minipage} &
    \begin{minipage}[t]{\linewidth}
    \begin{itemize}[leftmargin=*]
        \item The probability of certain paths in the KG may not be accurate.
        \item The lack of details on the classification model used.
    \end{itemize} 
    \vspace{1mm}
    \end{minipage} \\
  
    \hline 

    %==========================================%
Constructing knowledge graphs and their biomedical applications \cite{nicholson2020constructing} &
    
    \begin{minipage}[t]{\linewidth}
    \begin{itemize}[leftmargin=*]
        \item Examining the construction and application of biomedical knowledge graphs.
        \item Emphasizing how machine learning is transforming these processes.
    \end{itemize} 
    \vspace{1mm}
    \end{minipage} &
    \begin{minipage}[t]{\linewidth}
    \begin{itemize}[leftmargin=*]
        \item Discussion of knowledge graph construction, including manual curation and text mining.
        \item Review of representational learning techniques and their applications in biomedical fields.
    \end{itemize} 
    \vspace{1mm}
    \end{minipage}&
    \begin{minipage}[t]{\linewidth}
    \begin{itemize}[leftmargin=*]
        \item Need for advanced techniques to handle complex sentence structures.
        \item Limitations in current methods to represent diverse relationships in KGs.
        \item Scalability and memory limitations in matrix factorization techniques.
    \end{itemize} 
    \vspace{1mm}
    \end{minipage} \\
  
    \hline 

    %==========================================%
SMR: Medical knowledge graph embedding for safe medicine recommendation \cite{wang2017safe} &

Developing a framework to recommend safe medicines by leveraging a heterogeneous graph that integrates patient data, diseases, and medicines.&
    \begin{minipage}[t]{\linewidth}
    \begin{itemize}[leftmargin=*]
        \item Development of graph-based embedding models enabling the recommendation of newly emerged medicines effectively.
        \item A novel method to recommend safe medicines for new patients and minimizing potential adverse drug reactions.
        \item Introduction of the SMR framework as a new approach to the link prediction problem.

    \end{itemize} 
    \vspace{1mm}
    \end{minipage} &
    
    \begin{minipage}[t]{\linewidth}
    \begin{itemize}[leftmargin=*]
        \item Dealing with the challenge of recommending safe medicines, especially new ones, to patients.
        \item Minimizing potential adverse drug reactions in medicine recommendations is critical to patient safety.

    \end{itemize} 
    \vspace{1mm}
    \end{minipage} \\
  
    \hline 

%==========================================%
Patient centric approach for clinical trials: Current trend and new opportunities\cite{sharma2015patient} &

Exploring the shifting paradigm in clinical trials towards a more patient-centric model.&
Identifying new opportunities for the clinical research industry to adopt patient-centric approaches to accelerate drug development and improve trial outcomes.
 &
    
    \begin{minipage}[t]{\linewidth}
    \begin{itemize}[leftmargin=*]
        \item The complexity and rising costs of clinical research.
        \item Ensuring data transparency and building trust with patients participating in clinical trials.
    \end{itemize} 
    \vspace{1mm}
    \end{minipage} \\
  
    \hline

%==========================================%
Automatic diagnosis with efficient medical case searching based on evolving graphs\cite{wang2018automatic} &

Developing a method for automatic diagnosis by improving medical case searching using evolving graphs, which dynamically incorporate new medical cases and knowledge.&
    \begin{minipage}[t]{\linewidth}
    \begin{itemize}[leftmargin=*]
        \item Introduction of an evolving graph framework that integrates new medical cases and knowledge.
        \item A novel method for medical case searching that leverages the evolving graph structure.
        \item An optimization strategy for embedding learning in the heterogeneous graph.
    \end{itemize} 
    \vspace{1mm}
    \end{minipage} &
    
    \begin{minipage}[t]{\linewidth}
    \begin{itemize}[leftmargin=*]
        \item Handling the dynamic nature of medical knowledge and cases.
        \item Balancing the computational complexity of embedding learning in a continuously evolving graph structure.
        \item Scalability and maintaining high accuracy and efficiency as the graph expands.
    \end{itemize} 
    \vspace{1mm}
    \end{minipage} \\
  
    \hline 

    %==========================================%
    \end{tabularx}

}
\end{table*}

Although PCKGs have proven their value in areas such as recommending individual interventions, predicting disease before onset, and improving clinical trials, their utility goes far beyond these. It is increasingly becoming common to use KGs in other innovative healthcare applications, such as optimizing hospital workflows, tailoring patient engagement strategies, and even developing telemedicine platforms. With these innovative tools, healthcare can be revolutionized by providing a more holistic, integrated view of patient data, as well as new opportunities for research and treatment methods.

Having explored the diverse applications and use cases of PCKGs in various domains, we now turn our attention to the challenges and future directions in this field. This transition allows us to critically examine the current limitations and envision potential advancements that could further enhance the utility and effectiveness of PCKGs.

\section{Research Challenges \& Discussion}
\label{current_challenges}
PCKGs in healthcare are designed to provide a comprehensive, unified view of patient data by integrating information from various sources, including EHRs, medical literature, and patient-generated data. These graphs aim to support better clinical decision-making and personalized patient care by representing complex medical data in an interconnected format that is more accessible and actionable for healthcare providers.

The current state of research in PCKGs is focused on overcoming several key challenges to maximize their potential in healthcare. Ji et al. \cite{ji2020survey} discuss the difficulties in knowledge acquisition, completion, and temporal KG development, which are crucial for maintaining up-to-date and comprehensive patient profiles. Building on this foundation, Chen et al. \cite{chen2019robustly} highlight the need for robustness in PCKGs, particularly in addressing sample size limitations and unmeasured confounders, to extend models to larger patient visits. Moreover, Rastogi and Zaki \cite{rastogi2020personal} emphasize the importance of designing, building, and operationalizing PCKGs that are tailored to individual patients. However, a significant hurdle remains in the actual construction of PCKGs. As noted by Gyrard et al. \cite{gyrard2018personalized} and Cong et al. \cite{cong2018constructing} point out the complexities in constructing PCKGs, which are often time-consuming and heavily reliant on the quality of source data. A significant challenge in the application of traditional KG embedding methods to patient-centric healthcare applications is their struggle with structural sparsity. Hu et al.\cite{Hu2021} argue that conventional methods, such as TransE and ConvE, while adept at mapping entities and relationships into a vector space, falter due to their reliance solely on KG triplets, neglecting the rich auxiliary texts that describe entities. This limitation significantly limits the comprehensiveness and utility of KGs in capturing detailed patient information and medical knowledge, indicating a key challenge in leveraging KGs for complex healthcare applications.

Data quality and standardization remain significant challenges, as heterogeneous data structures, poor data quality, and varying medical standards complicate the integration of data into a coherent KG Zhang et al.\cite{zhang2020hkgb}. The integration of PCKGs into clinical workflows also presents challenges, as it requires the development of systems that complement healthcare providers' routines without causing disruptions \cite{gortzis2008tracing}. Additionally, scalability is another concern, as PKGs must be able to incorporate an ever-increasing amount of data from diverse sources, including genomic information and patient lifestyle data \cite{wang2018automatic}. Patient data privacy is a critical concern, particularly in the context of utilizing PCKGs, which involve handling sensitive personal health information and require strict privacy controls to safeguard patient confidentiality. The use of patients' data for various purposes, such as consultations, research, and emergency, poses a significant challenge for authorization systems, emphasizing the need for robust privacy protection \cite{Al-Zubaidie.2019}\cite{sharma2022privacy}. Furthermore, real-time data analysis within PCKGs is technologically demanding, requiring advanced computational methods to process and analyze data promptly for it to be clinically relevant \cite{hooshafza2021modelling}. 

In light of these challenges, it becomes clear that continued technological and methodological
innovations are necessary to enhance the predictive capabilities of PCKGs. The use of advanced
analytics, machine learning, and semantic web technologies could be key. Additionally, as PCKGs
become more integrated into healthcare delivery, it becomes increasingly important to address
regulatory and ethical considerations. This requires collaboration among computer scientists,
healthcare professionals, and policymakers to align the development of PCKGs with broader
healthcare objectives.

\section{Conclusion \& Future Directions}
\label{conclusion}
This literature review of PCKGs explores their development, evaluation, processing techniques, applications, challenges, and prospects. PCKGs represent a field in healthcare informatics that aims to revolutionize personalized patient care by integrating and synthesizing diverse healthcare data sources. The review highlights the current state-of-the-art methodologies for constructing and evaluating PCKGs, emphasizing the importance of qualitative and quantitative approaches to assess their effectiveness in healthcare settings. In addition to construction and evaluation, the review delves into innovative processing techniques such as reasoning, semantic search, and inference. These techniques significantly enhance the accuracy and efficiency of PCKGs, ultimately improving patient-centered care. Furthermore, exploring different applications of PCKGs in healthcare—including disease prediction, personalized treatment recommendations, and advancements in clinical trials—reveals their potential to transform healthcare through personalized and predictive medicine.

Future directions in PCKG research include leveraging advanced analytics and machine learning to improve predictive capabilities, which could lead to more accurate and timely interventions \cite{wu2021construct}. In addition, semantic web technologies are also predicted to play a significant role in enhancing the accessibility and utility of PCKGs \cite{huang2017constructing}. Building on this momentum, personalized medicine emerges as a promising area where PCKGs can make a substantial impact. By linking genomic data with clinical outcomes, treatments can be tailored to individual patients, offering a more personalized approach to healthcare \cite{primekg}. To further this advancement, methodological innovations, such as new algorithms for data harmonization and user interface design, are crucial. These innovations are needed to address current challenges and facilitate the broader adoption of PCKGs in clinical practice \cite{peng23challenges}. Simultaneously, as PCKGs become more integrated into healthcare delivery, regulatory and ethical considerations gain prominence. These aspects are critical to ensuring that the deployment of PCKGs adheres to the highest standards of patient care and data management. Therefore, cross-disciplinary collaboration becomes essential for advancing PCKG technology. This involves computer scientists, healthcare professionals, and policy-makers working together to ensure that the development of PCKGs aligns with broader healthcare objectives and respects ethical guidlines\cite{solanki2022ethics}\cite{maxhelaku2022exploit}.

PCKGs represent a significant advancement in healthcare informatics. They can transform patient outcomes by enabling early detection and personalized treatment plans. Despite the challenges involved, the future outlook for PCKGs is promising as they have the potential to improve patient care and healthcare delivery significantly. Ongoing research efforts and interdisciplinary collaboration will be crucial in fully realizing their novel impact on healthcare.

\section*{Acknowledgments}
This work was supported by PATENT Lab (Predictive Analytics and TEchnology iNTegration Laboratory) at the Department of Computer Science and Engineering, Mississippi State University.

\section*{Conflict of Interest Statement} 

The authors declare that the research was conducted in the absence of any commercial or financial relationships that could be construed as a potential conflict of interest.

\section*{Author Contributions}

All authors contributed to producing the paper. All authors
contributed to the article and approved the submitted version.

%\section*{References}

\bibliographystyle{unsrt}
%\raggedbottom
\bibliography{9-Refs}

\end{document}